\documentclass[11pt, a4paper, logo, copyright]{deepmind}

\usepackage{amsmath}
\usepackage{amssymb}
\usepackage{amssymb}
\usepackage{amsthm}
\usepackage{graphicx}
\usepackage{subcaption}
\usepackage[authoryear, sort&compress, round]{natbib}
\usepackage{amsmath}
\DeclareMathOperator*{\argmax}{arg\,max}

\theoremstyle{definition}
\newtheorem{definition}{Definition}[section]

\title{The Hydra Effect: Emergent Self-repair in Language Model Computations}

\author[1]{Thomas McGrath}
\author[1]{Matthew Rahtz}
\author[1]{J\'{a}nos Kram\'{a}r}
\author[1]{Vladimir Mikulik}
\author[1]{Shane Legg}

\affil[1]{Google DeepMind}

\begin{abstract}
    We investigate the internal structure of language model computations using causal analysis and demonstrate two motifs: (1) a form of adaptive computation where ablations of one attention layer of a language model cause another layer to compensate (which we term the Hydra effect) and (2) a counterbalancing function of late MLP layers that act to downregulate the maximum-likelihood token. Our ablation studies demonstrate that language model layers are typically relatively loosely coupled (ablations to one layer only affect a small number of downstream layers). Surprisingly, these effects occur even in language models trained without any form of dropout. We analyse these effects in the context of factual recall and consider their implications for circuit-level attribution in language models.
\end{abstract}

\begin{document}

\maketitle

\begin{figure}[h!]
    \centering
    \includegraphics[width=\textwidth]{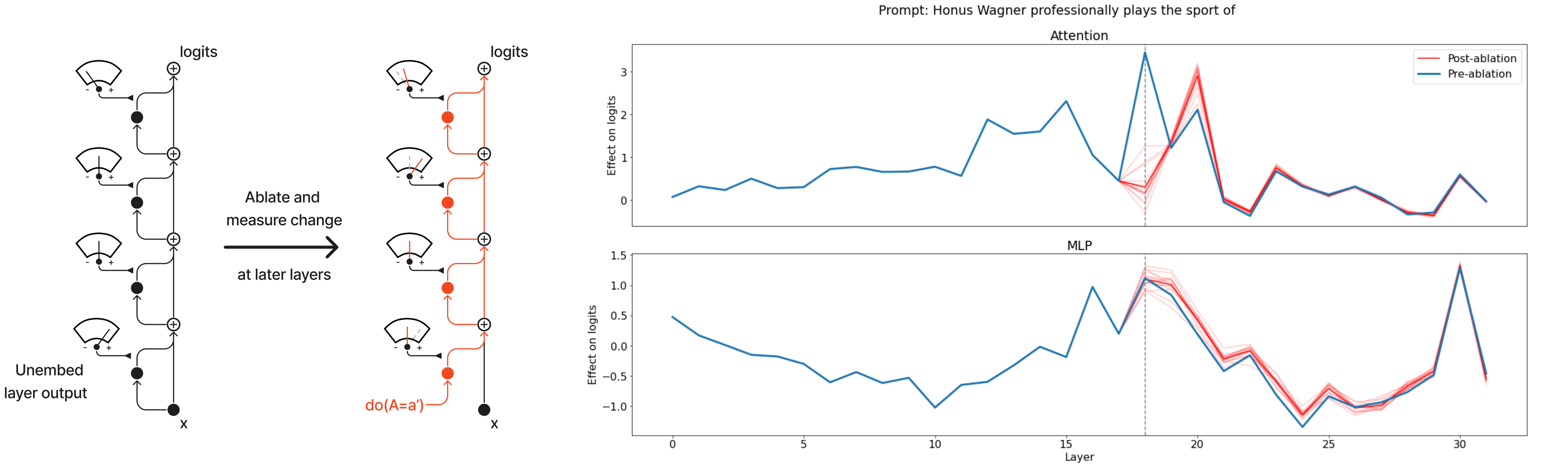}
    \caption{Diagram of our protocol for investigating network self-repair and illustrative results. The blue line indicates the effect on output logits for each layer for the maximum-likelihood continuation of the prompt shown in the title. Faint red lines show direct effects following ablation of at a single layer indicated by dashed vertical line (attention layer 18 in this case) using patches from different prompts and the solid red line indicates the mean across patches. See Section~\ref{s:hydra-explanation} for details.}
    \label{fig:hydra-effect_examples}
\end{figure}

\section{Introduction}
Ablation studies are a vital tool in our attempts to understand the internal computations of neural networks: by ablating components of a trained network at inference time and studying the downstream effects of these ablations we hope to be able to map the network's computational structure and attribute responsibility among different components. In order to interpret the results of interventions on neural networks we need to understand how network computations respond to the types of interventions we typically perform. A natural expectation is that ablating important components will substantially degrade model performance~\citep{morcos2018importance} and may cause cascading failures that break the network. We demonstrate that the situation in large language models (LLMs) is substantially more complex: LLMs exhibit not just redundancy but actively self-repairing computations. When one layer of attention heads is ablated, another later layer appears to take over its function. We call this the Hydra effect: when one set of heads is cut off, other heads grow in importance\footnote{Although evocative, we recognise that the use of the name `Hydra' is not completely mythologically accurate: sometimes only one head grows in importance, and as we show in Section~\ref{s:quantifying-hydra} the total effect decreases on average, in contrast with the behaviour of the mythological Hydra.}. We present these results in Section~\ref{s:hydra-explanation}.
\newline\newline
The Hydra effect (referred to in~\citep{wang2022interpretability} as backup behaviour) complicates our understanding of what it means for a network component to be important because two natural-seeming measures of importance (unembedding and ablation-based measures) become much less correlated than we would na\"ively expect. These impact measurements correspond to the direct and total effect, which we introduce in a short, self-contained primer on causal inference in neural networks in Section~\ref{s:nns_as_causal_models} before performing a more comprehensive quantitative analysis of the Hydra effect in Section~\ref{s:quantifying-hydra}. Finally we discuss related work in Section~\ref{s:related-work} and close by hypothesising possible causes for our findings and discuss their implications for future causal analyses of language model computations in Section~\ref{s:conclusion}.

\section{Self-repair and the Hydra effect}\label{s:hydra-explanation}
\subsection{The Transformer architecture for autoregressive language modelling}\label{ss:transformer-architecture}
We want to analyse the computational structure of large autoregressive language models with an decoder-only Transformer architecture. In this work we use a 7 billion parameter model from the Chinchilla family (meaning that the architecture and training setup is identical to that proposed in ~\citep{hoffmann2022training} but the model is approximately 7 billion parameters and the training dataset size is scaled down appropriately). An autoregressive language model maps a sequence of input tokens $x_{\leq t}=(x_1, \ldots, x_t)$ of length $t$ to a probability distribution over the next token $x_{t+1}$ using a function $f_\theta$
\begin{align}
    p(x_{t+1}|x_{\leq t}) &= f_{\theta}(x_{\leq t})\\
    &= \mathrm{softmax}\left(\pi_t(x_{\leq t})\right),
\end{align}
where the pre-softmax values $\pi$ are called the logits. The function $f_{\theta}$ is a standard Transformer architecture comprised of $L$ layers
\begin{align}\label{eq:transformer-eqns}
    \pi_t &= \mathrm{RMSNorm}(z^L_t)W_U\\
    z^l_t &= z^{l-1}_t + a^l_t + m^l_t\\
    a^l_t &= \mathrm{Attn}(z^{l-1}_{\leq t})\\
    m^l_t &= \mathrm{MLP}(z^{l-1}_t),\label{eq:transformer-eqns-final}
\end{align}
where $\mathrm{RMSNorm}(\cdot)$ is an RMSNorm normalisation layer, $W_U$ an unembedding matrix $\mathrm{Attn}(\cdot)$ an attention layer~\citep{bahdanau2014neural, vaswani2017attention} and $\mathrm{MLP}(\cdot)$ a two-layer perceptron. The dependence of these functions on the model parameters $\theta$ is left implicit. In common with much of the literature on mechanistic interpretability (e.g.~\cite{elhage2021mathematical}) we refer to the series of residual activations $z^l_i$, $i=1,\ldots, t$ as the residual stream. For more details on the Transformer architecture in language modelling and the specific details of Chinchilla language models see~\citep{phuong2022formal, hoffmann2022training}. As an additional notational shorthand we will typically denote the dependence of network activations on inputs by skipping the repeated function composition and simply writing $z^l_t(x_\leq t)$ (or $a^l_t(x_{\leq t}$, $m^l_t(x_{\leq t})$) to denote the activations at layer $l$, position $t$ due to input string $x_{\leq t}$ rather than writing the full recursive function composition implied by Equations~\ref{eq:transformer-eqns}-\ref{eq:transformer-eqns-final}.

\subsection{Using the Counterfact dataset to elicit factual recall}
The Counterfact dataset, introduced in~\citep{wang2022interpretability}, is a collection of factual statements originally intended to evaluate the efficacy of model editing. The dataset comprises a series of prompts formed by combining tuples of subject, relation, and object $(s, r, o^*, o^c)$, where $s$ is the subject (e.g. Honus Wagner), $r$ the relation (e.g. ``professionally plays the sport of''), $o^*$ the true object (e.g. baseball), and $o^c$ some counterfactual claim that makes sense in context. We are interested in the way that language models store and relate factual knowledge so we only use the concatentation of $s$ and $r$ to form our prompts. This produces prompts whose completion requires factual knowledge that Chinchilla 7B can complete answer correctly.

\subsection{Measuring importance by unembedding}
One way to assess the effect of a neural network layer is to attempt to map its outputs onto the network output logits. We discuss approaches that use a learned probe in Section~\ref{s:related-work}, however in this work we  use the model's own unembedding mechanism $u$ to compute effects. This approach, often referred to as the logit lens, was introduced in~\citep{nostalgebraist2020logit} and has also been used in subsequent interpretability research~\citep{dar2022analyzing, geva2022lm}. The GPT-2 model used in the original logit lens analysis had a LayerNorm~\citep{ba2016layer} prior to unembedding, however in the Chinchilla model we analyse the unembedding function is an RMSNorm~\citep{zhang2019root} followed by an unembedding matrix $W_U$:
\begin{equation}
    u(z^l) = \mathrm{RMSNorm}(z^l)W_U.
\end{equation}
RMSNorm is a simplification of LayerNorm referred to as RMSNorm~\cite{zhang2019root} which dispenses with centring and the learned bias term:
\begin{equation}
    \mathrm{RMSNorm}(z) = \frac{z}{\sigma(z)}G;\quad \sigma(z) = \sqrt{\frac{1}{d}\sum^{d}_{i=1}z_i^2}
\end{equation}
where $z_i$ is the $i$-th element of the vector $z$ and $G$ is a learned diagonal matrix. In the remainder of this paper, unembedding refers to computing the logit lens distribution $\tilde{\pi}_t = u(z^l_t)=\mathrm{RMSNorm}(z^l_t)W_U$ using the model's final RMSNorm layer.
\subsubsection{Impact metric}\label{ss:unembed-impact}
Our unembedding-based impact measurement $\Delta_{\mathrm{unembed}}$ will be measured in terms of model logits over the vocabulary $V$. Whenever we deal with logits, either from inference by the complete model or from unembedding, we will first centre the logits:
\begin{equation}\label{eq:logit-centring}
    \hat{\pi}_t = \pi_t - \mu_\pi;\quad\mu_\pi = \frac{1}{V}\sum^V_{j=1}[\pi_t]_j,
\end{equation}
where $[\pi_t]_j$ indicates the logit of the $j$-th vocabulary element. We measure the impact of an attention layer $a^l$ on the logit of the maximum-likelihood token at the final token position $t$:
\begin{equation}\label{eq:ablation-impact}
    \Delta_{\mathrm{unembed}, l} = \hat{u}(a^l_t)_i;\quad i = \argmax_j[\pi_t]_j,
\end{equation}
where $\hat{u}$ denotes the centred version of the logits obtained by unembedding as in Equation~\ref{eq:logit-centring}. We also repeat the procedure for the MLP layers $m^l$, $l=1,\ldots,L$. We fix the RMSNorm normalisation factor $\sigma$ to the value attained in the forward pass, i.e. $\sigma = \sigma(z^L_t)$ where $L$ denotes the final layer of the network. Fixing the normalisation factor means that the effect of each layer output on the logits is linear, and corresponds to the unembedding that is actually carried out during the model's forward pass. As we will show in Section~\ref{s:nns_as_causal_models}, unembedding in this way corresponds to the direct effect from causal inference.

\subsection{Measuring importance by ablation}
An alternative approach to investigating a layer's function is to ablate it by replacing its output with some other value, for example replacing $a^l_t$ with an identically-shaped vector of zeros during network inference would ablate $a^l_t$. This approach has appealing parallels to standard methodologies in neuroscience and has been suggested as a gold standard for evidence in interpretability research~\citep{leavitt2020towards}. We naturally expect that ablating components that are important for a given input will lead to degraded performance on that input (indeed this seems like a reasonable definition of importance) and if a network's internal mechanisms generalise then the ablation will also lead to degraded performance on other similar inputs.
\subsubsection{Intervention notation \& impact metric}\label{ss:ablation-impact}
In keeping with standard notation in causality~\citep{glymour2016causal, pearl2009causality} we indicate replacing one set of activations $a^l_t$ with another using the $\mathrm{do}(\cdot)$ operation. Here we need to introduce some additional notation: we refer to specific nodes in the model's compute graph using capital letters and their actual realisation on a set of inputs with lowercase. For example: $A^l_t$ refers to the attention output at layer $l$ and position $t$, whereas  $a^l_t(x_{\leq t})$ refers to the value of this vector when the model is evaluated on inputs $x_{\leq t}$ (when the dependence on inputs is omitted for brevity it should be either clear from context or not relevant). If we have inputs $x_{\leq t}$ that result in logits $\pi_t(x_{\leq t})$ we would write the value of $\pi_t(x_{\leq t})$ following an intervention on $A^l_t$ as
\begin{equation}
    \pi_t\left(x_{\leq t} | \mathrm{do}(A^l_t = \tilde{a}^l_t)\right) = \pi_t\left(x_{\leq t} | \mathrm{do}\left(A^l_t = a^l_t(x'_{\leq t})\right)\right)
\end{equation}
for some alternative input $x'_{\leq t}$ (see Appendix~\ref{s:appendix-intervention-distribution} for details of how alternative inputs can be chosen). As with the unembedding impact measure $\Delta_{\mathrm{unembed}}$ we measure the impact of ablation $\Delta_{\mathrm{ablate}}$ on the centred logit $\hat{\pi}$ of the maximum-likelihood token $i$ for a given input $x_{\leq t}$ (see Section~\ref{ss:unembed-impact}). To compute $\Delta_{\mathrm{ablate}, l}$ of attention layer $l$, token position $t$ we instead compare the centred logit of the maximum-likelihood token $i$ before and after intervention:
\begin{equation}\label{eq:ablate-defn}
    \Delta_{\mathrm{ablate}, l} = \left[\hat{\pi_t}(x_{\leq t} | \mathrm{do}(A^l_t = \tilde{a}^l_t)) - \hat{\pi_t}(x_{\leq t}) \right]_i.
\end{equation}
As we will show in Section~\ref{s:nns_as_causal_models}, Equation~\ref{eq:ablate-defn} corresponds to measuring the total effect of $A^l_t$ in the context $x_{\leq t}$. We used resample ablation~\citep{chan2022causal} with patches from 15 alternative prompts from the dataset to provide ablation activations $\tilde{a}^l_t$ (see Appendix~\ref{s:appendix-intervention-distribution} for more details).

\subsection{Ablation-based and unembedding-based importance measures disagree in most layers}
We calculate the importance of each layer according to both ablation and unembedding-based measures of importance for all layers of Chinchilla 7B for all prompts in the Counterfact dataset. For each prompt we calculate $\Delta_{\mathrm{unembed}, l}$ and $\Delta_{\mathrm{ablate}, l}$ at the final token position for every attention and MLP layer.

\begin{figure}[ht!]
    \centering
    \begin{subfigure}{0.45\textwidth}
    \centering
    \includegraphics[width=\textwidth]{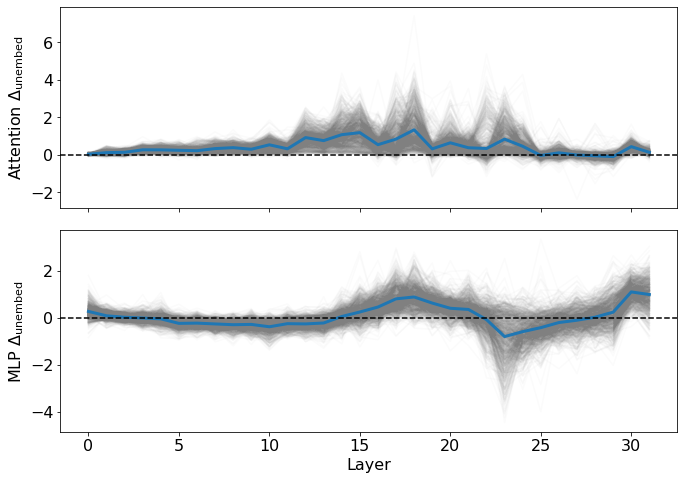}
    \caption{}
    \end{subfigure}
    \begin{subfigure}{0.45\textwidth}
    \centering
    \includegraphics[width=\textwidth]{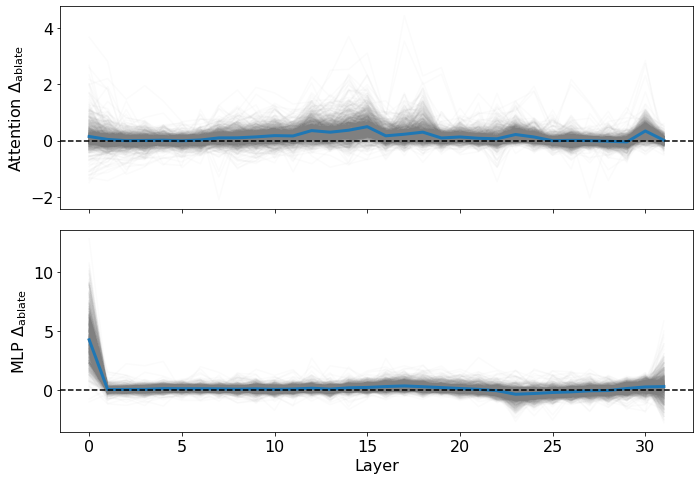}
    \caption{}
    \end{subfigure}
    
    \begin{subfigure}{0.45\textwidth}
    \centering
    \includegraphics[width=\textwidth]{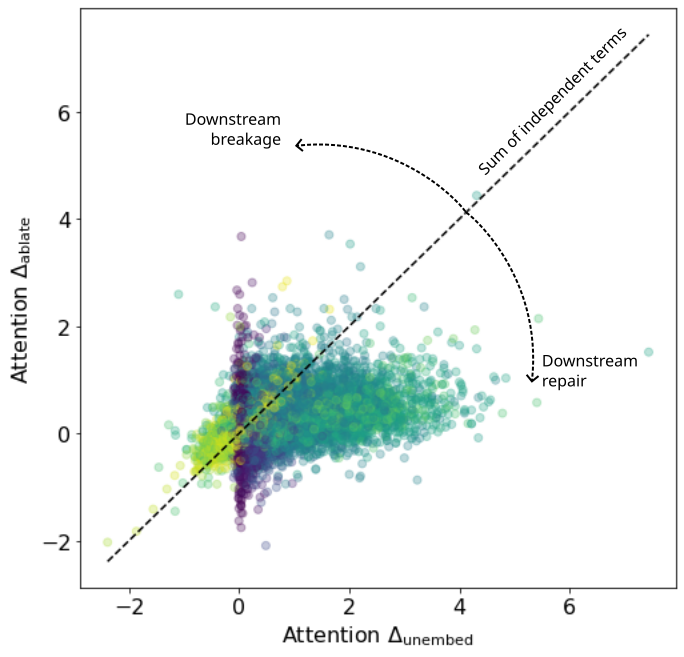}
    \caption{}
    \end{subfigure}
    \begin{subfigure}{0.45\textwidth}
    \centering
    \includegraphics[width=\textwidth]{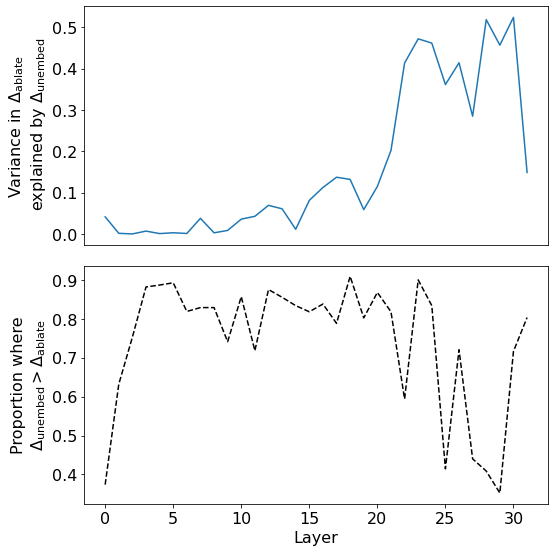}
    \caption{}
    \end{subfigure}
    \caption{Measurements of (a) unembedding-based and (b) ablation-based impact for MLP and attention layers for a 7B parameter language model on the Counterfact dataset. Grey lines indicate per-prompt data, blue line indicates average across prompts. (c) Comparison of unembedding- and ablation-based impact measures across all layers showing low correlation and $\Delta_{\mathrm{unembed}} > \Delta_{\mathrm{ablate}}$ for most prompts and layers, contrary to expectations. (d) Quantification of correlation between $\Delta_{\mathrm{unembed}}$ and $\Delta_{\mathrm{ablate}}$ and proportion of prompts where $\Delta_{\mathrm{unembed}} > \Delta_{\mathrm{ablate}}$ across layers.}
    \label{fig:overall-effects}
\end{figure}
These results are shown in Figure~\ref{fig:overall-effects}, which shows average and per-prompt results, demonstrating a substantial disagreement between $\Delta_{\mathrm{unembed}}$ and $\Delta_{\mathrm{ablate}}$. This is surprising as we would expect that ablating a layer not only destroys its impact on the logits from unembedding but potentially also breaks further downstream network components, so we would expect that the ablation-based measure would be greater than or equal to the unembedding measure. As Figure~\ref{fig:overall-effects} shows, this is far from the case. We now demonstrate that the lack of correlation between ablation and unembedding measures of component importance at all but the late layers of the network is due to downstream changes in layer outputs that counteract the effect of ablations.

\subsubsection{Methodology}
In order to understand the mechanism behind the difference in results from ablation and unembedding methods we propose a simple methodology that allows us to localise changes in network computation. First, for each input $x_{\leq t}$ we compute attention and MLP outputs $a^l_{t}(x_{\leq t})$ and $m^l_{t}(x_{\leq t})$ for all layers $l$. We refer to this as the clean run, matching the terminology from~\citep{meng2022locating}. We then compute the unembedding-based impact $\Delta_{\mathrm{unembed}, l}$ as defined in Section~\ref{ss:unembed-impact}. This gives us a per-layer measure of the impact of each layer on the maximum-likelihood logit in the clean run. We then ablate a specific attention or MLP layer $k$ using resample ablation (see Appendix~\ref{s:appendix-intervention-distribution}). We refer to this as the layer $k$ ablation run. We now compute $\Delta_{\mathrm{unembed}, l}$ for each layer $l$ and layer $k$ ablation run. We denote a specific unembedding layer $l$ and ablation layer $k$ by $\tilde{\Delta}^{k}_{\mathrm{unembed}, l}$:
\begin{equation}
    \tilde{\Delta}^k_{\mathrm{unembed}, l} = u\left(a^l_t\,|\,\mathrm{do}(A^k_t = \tilde{a}^k_t)\right).
\end{equation}
Because the transformer network is a feedforward network, if a readout layer $l$ is not causally downstream of the ablation layer $k$ then $\tilde{\Delta}^{k}_{\mathrm{unembed}, l} = \Delta_{\mathrm{unembed}, l}$. If $k=l$ then $\tilde{\Delta}^{k}_{\mathrm{unembed}, l}\approx 0$ because the output of that layer will be ablated (the approximate equality is due to the fact that the resample ablation is stochastic and so may not fully zero out the centred logit of the maximum-likelihood token).
\newline\newline
This methodology, depicted in Figure~\ref{fig:hydra-effect_examples}, measures the impact of each layer on the maximum-likelihood token before and after ablation. This allows us to determine how individual layers react to a given ablation.


\subsubsection{Results}
\begin{figure}
    \centering
    
    \begin{subfigure}{0.6\textwidth}
        \includegraphics[width=\textwidth]{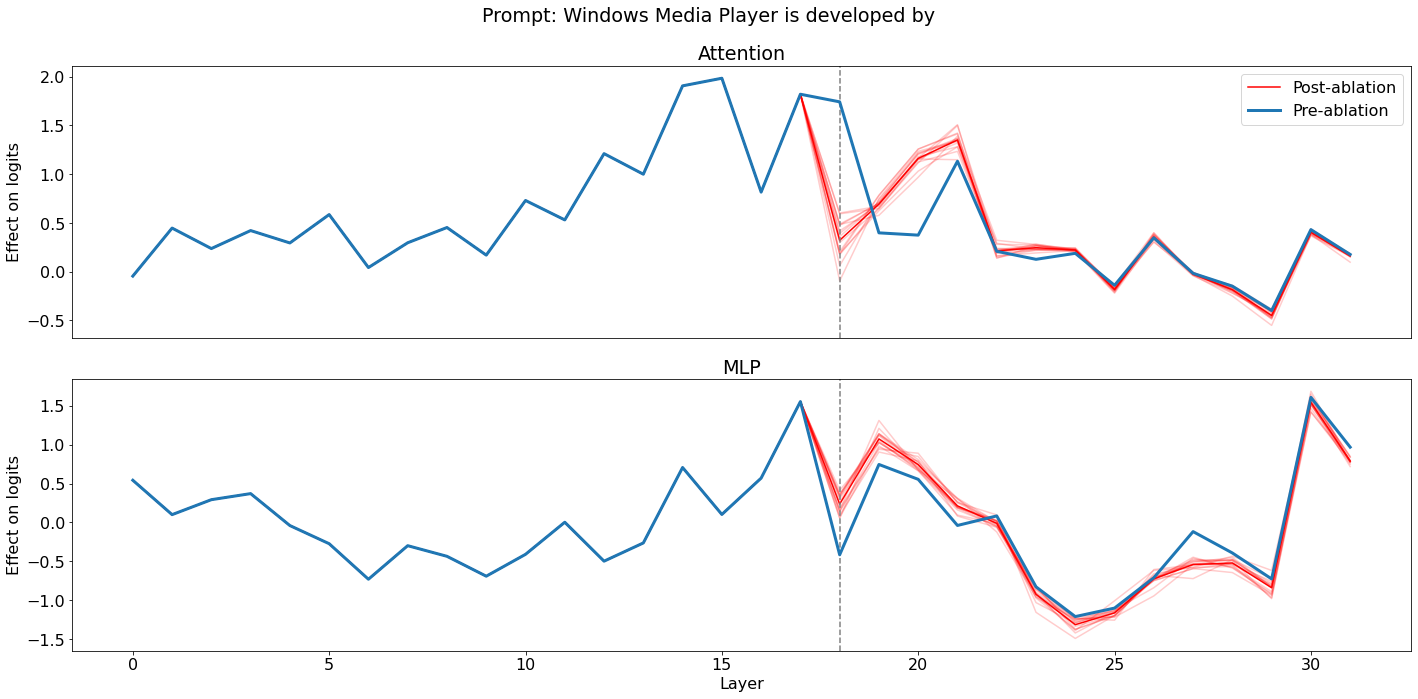}
    \end{subfigure}
    
    \begin{subfigure}{0.6\textwidth}
        \includegraphics[width=\textwidth]{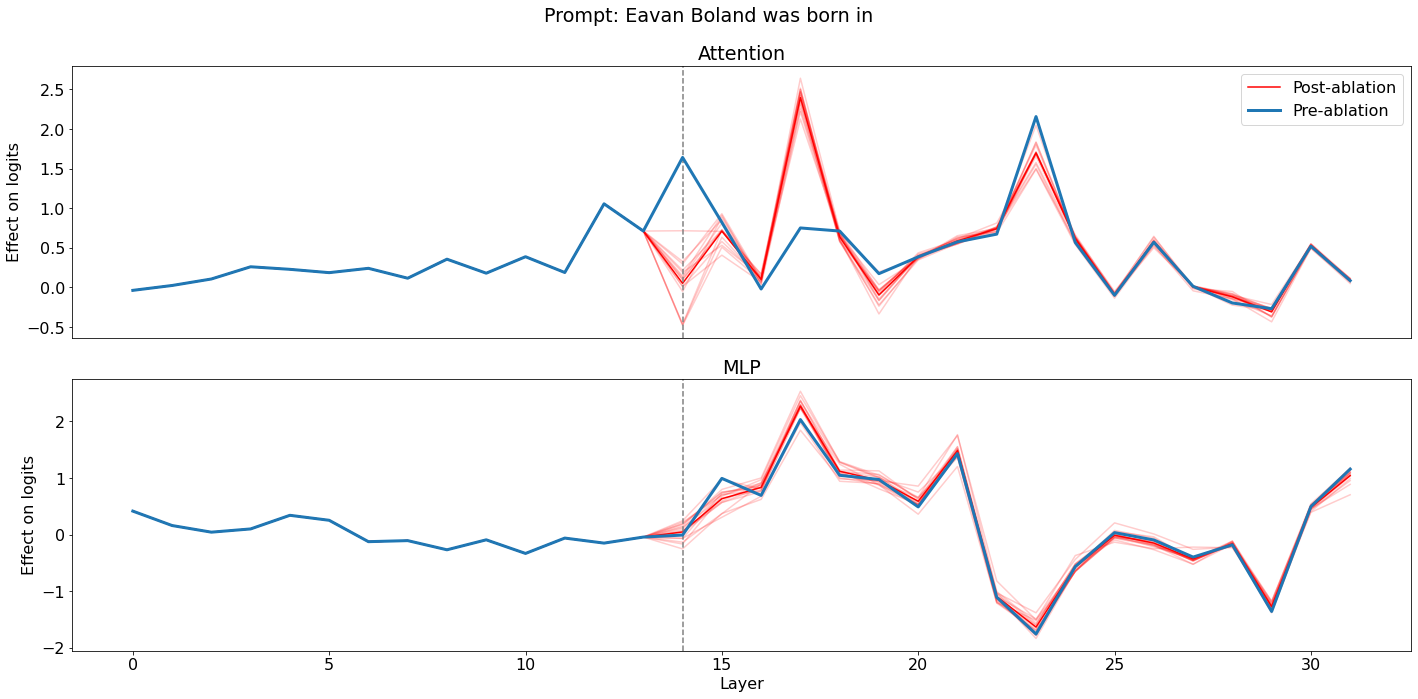}
    \end{subfigure}
    
    \caption{Example results of self-healing computations in Chinchilla 7B showing per-layer impact on final logits before and after ablation.}
    \label{fig:hydra-effect_examples-2}
\end{figure}
Sample results for attention ablation are shown in Figure~\ref{fig:hydra-effect_examples} and Figure~\ref{fig:hydra-effect_examples-2} showing ablations of layers with high $\Delta_{\mathrm{unembed}}$. Several observations stand out from this data:
\paragraph{Resample ablations work but are noisy:} Resample ablations successfully reduce $\Delta_{\mathrm{unembed}}$ of the ablated layer to approximately zero when averaged across patches, although there is substantial variance between the effect of each patch. This demonstrates that resample ablations can provide true ablations but require a reasonably large sample size to prevent excessive noise.
\paragraph{The Hydra effect occurs in networks trained without dropout:} In both cases an attention layer downstream of the ablated layer (layer 20 in the examples shown) substantially \textit{increases} its impact in the ablated network compared to the intact network, i.e. $\tilde{\Delta}^m_{\mathrm{unembed}, l} > \Delta_{\mathrm{unembed}, l}$ for some unembedding layer $l > m$. These are examples of the Hydra effect: we cut off some attention heads, but others grow their effect to (partially) compensate. The Chinchilla-like language model we study here was trained entirely without dropout, stochastic depth, or layer dropout.
\paragraph{Downstream effects of ablation on attention layers are localised:} The effects of attention ablation on downstream layers are localised: apart from the subsequent attention layer involved in the Hydra effect, other attention layers have $\Delta_{\mathrm{unembed}}$ almost entirely unchanged in the ablated network. This does not necessarily imply that the features that would have been created by the ablated layer are unused in later layers, however (although it is consistent with this hypothesis). It is possible that the Hydra effect is not just compensating for the impact of the ablated layer on logits, but is also replacing the missing features that the ablated layer would have provided. More detailed multiple ablation studies would be needed to distinguish between these hypotheses.
\paragraph{Downstream MLPs may perform erasure/memory management:} The shape of the unembedding impact curve across layers for the MLP layers remains very similar, but the impact of many of the layers is attenuated. This suggests that these MLPs are performing an erasure/memory-management role: when the attention layer has a high positive impact they have a high negative impact and when the attention layer's $\Delta_{\mathrm{unembed}}$ is reduced theirs is similarly attenuated.
\newline\newline
Although these observations are surprising, the evidence we have presented here is largely anecdotal. Before expanding our analysis to the full Counterfact dataset and analysing all layers in Section~\ref{s:quantifying-hydra} we first introduce basic tools of causal inference and reframe our analyses in terms of concepts from causality in Section~\ref{s:nns_as_causal_models}.

\section{Neural networks as causal models: the compute graph is the causal graph}\label{s:nns_as_causal_models}
This section introduces structural causal models, causal graphs and the idea of interventions. The central idea of this section is that we can consider the internal structure of neural networks as structural causal models and use tools from causality to analyse their internal workings. One thing we are \textit{not} claiming is that neural networks are naturally forming causal models of their training data, or that networks learn to perform causal reasoning - the degree to which this happens in practice or forms the basis for successful generalisation is still not well understood.
\paragraph{Causal models}
\begin{definition}[Causal model]
A structural causal model $M$ is a tuple $M = \langle U, V, F, P(u) \rangle$ where:
\begin{enumerate}
    \item $U$ and $V$ are sets of variables called the exogenous and endogenous variables respectively. We follow the standard convention of writing a random variable in uppercase and a specific realisation of that variable in lowercase, so $u$ is a realisation of the exogenous variable $U$.
    \item $F$ is a set of deterministic functions where $f_i \in F$ determines the value of $V_i$ from a subset of variables $\mathrm{Pa}_i \in U \cup V \setminus\left\{V_i\right\}$, i.e. $v_i = f_i(\mathrm{pa}_i)$.
    \item $P(u)$ is a probability distribution over exogenous variables $U$.
\end{enumerate}
\end{definition}
\begin{figure}[t!]
    \centering
    \includegraphics[width=0.7\textwidth]{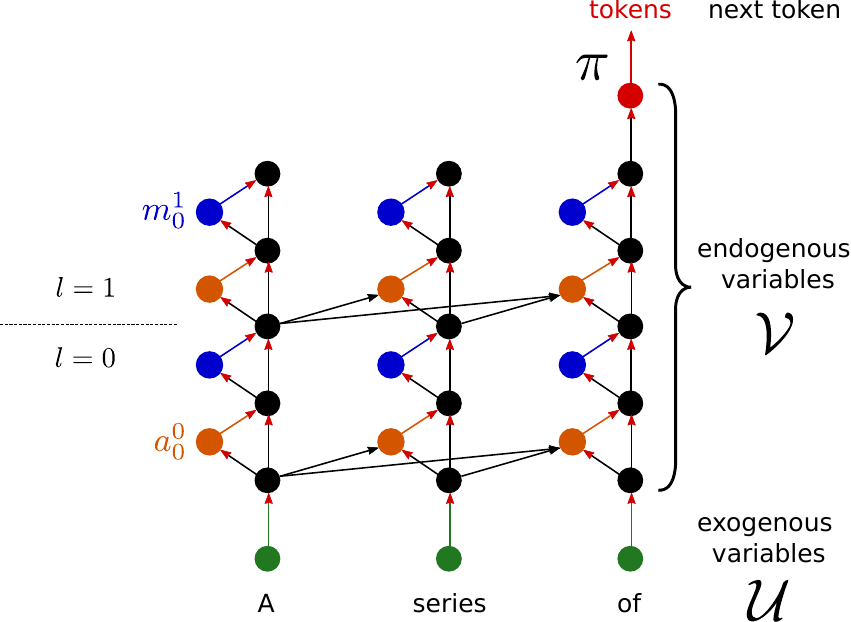}
    \caption{Viewing a transformer network as a structural causal model.}
    \label{fig:nn_as_scm}
\end{figure}

In a structural causal model the exogenous variables $U$ represent randomly-set external conditions (the `context') and the endogenous variables $V$ follow deterministically from the context $u$. Given a setting of exogenous variables $U=u$, the value of an endogenous variable $V_i$ is fully determined. We write the value $V_i$ takes in this situation as $V_i(u)$ to reflect this rather than write the full set of functions relating $V_i$ to $U$ for conciseness and generality.

\paragraph{}We can visualise structural causal models as directed graphs where an edge $X_i \to X_j$ (for $X_i,\,X_j\in U \cup V$) exists if $X_i\in\mathrm{pa}_i$, i.e. if $X_i$ is one of the variables that directly determines the value of $X_j$. The important factor distinguishing $U$ from $V$ is that variables in $U$ have no in-edges: there is nothing apart from the probability distribution $P(u)$ that determines their value. Variables in $V$, on the other hand, can eventually trace their ancestry back to one or more values in $U$ (although in large models it may take many edges to do so). 

\begin{figure}[t!]
    \centering
    \includegraphics[width=0.9\textwidth]{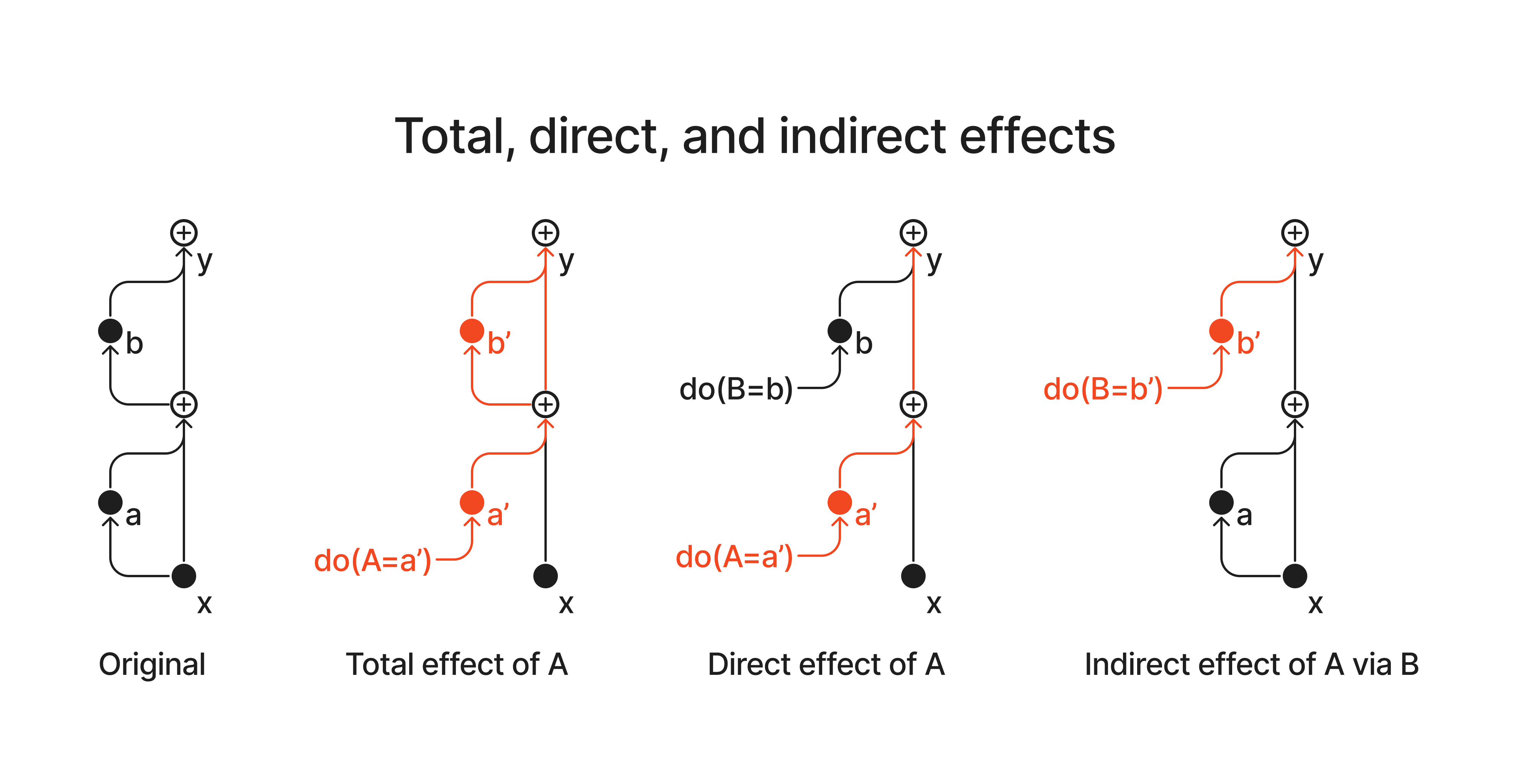}
    \caption{Interventions and types of effect illustrated with a two-layer residual network example. The total effect involves intervening on a node and allowing the changes due to that intervention to propagate arbitrarily, whereas the direct effect only allows those changes to propagate via the direct path between intervention and output. The indirect effect is the complement of the direct effect: effects are allowed to flow via all but the direct path.}
    \label{fig:interventions}
\end{figure}

\paragraph{Interventions \& counterfactuals in causal models}
In a causal model an intervention $\mathrm{do}(Z=z')$ alters the value of an endogenous variable $Z$ from whatever value it would have taken based on its parents $z = f_Z(\mathrm{pa}_Z)$ to the new value $z'$. This entails a change in the causal model $M$ to a new intervened-upon model $M_Z$ where $M_Z$ is identical to $M$ except that $f_Z(\mathrm{pa}_Z) = z'$ regardless of the values taken by $\mathrm{pa}_Z$: the function $f_Z$ has been replaced by a constant function returning $z'$. In terms of the causal graph this leads to a removal of in-edges to $Z$ (as it no longer depends on other elements of the graph) and the addition of a new intervention edge (see Figure~\ref{fig:interventions} for some examples). We express this intervention using the $\mathrm{do}$-operator \citep{pearl2009causality}, where $Y(u \,|\,\mathrm{do}(Z=z'))$ denotes the value that $Y$ takes in the modified model $M_Z$ given the context $u$.

\paragraph{Treating language models as causal models} Given these preliminaries the correspondence between neural networks and structural causal models is hopefully clear: the input data $x$ correspond to the exogenous variables $U$ and the network's activations ($z$, $a$, $m$) and outputs $\pi$ correspond to the endogenous variables $V$. In autoregressive transformer-based language models causal masking ensures that activations at input position $t_1$ are never causal parents of activations at input position $t_2$ if $t_1 > t_2$. For the remainder of this work we will assume a standard Transformer architecture, as shown in Figure~\ref{fig:nn_as_scm}. Intervening on neural networks is identical to intervening in a SCM: we set the nodes to their intervention values and propagate the effects of this intervention forwards. Although for concreteness we focus on language modelling using transformers, the general framework of causal analysis is applicable to deterministic feedforward neural networks more generally - all that changes is the structure of the graph. Stochastic networks such as VAEs or diffusion models can be incorporated by adding additional exogenous nodes corresponding to sources of noise.

\subsection{Direct, indirect, and total effects}\label{ss:simple-effect-types}
We now have the notation and concepts needed to define and compute several important types of effects in causal graphs: the total, direct, and indirect effects. When we define the different types of effect in a causal model we will always consider the effects in a given context (i.e. a given setting of exogenous variables $u$).

\begin{definition}[Total effect]
The total effect of altering the value of a variable $Z$ from $Z=z$ to $Z=z'$ on another variable $Y$ in a context $u$ is given by
\begin{equation}\label{eq:total-effect}
    TE(z\to z', Y, u) = Y(u | \mathrm{do}(Z=z')) - Y(u | \mathrm{do}(Z=z)).
\end{equation}
\end{definition}
\paragraph{Total effect corresponds to ablation-based impact} If we choose $Y = \hat{\pi}_t$ (i.e. the random variable we measure the effect of the intervention on is the centred logit of the maximum-likelihood token $i$ at the final token position $t$) and the exogenous variables are the prompt $u=x_{\leq t}$ then we the total effect of an intervention to change activations $A^l_t$ from their `natural' $a^l_t=a^l_t(x_{\leq t})$ to an intervened-upon value $\tilde{a}^l_t = a^l_t(x'_{\leq t})$ is
\begin{align}
    TE(a^l_t\to\tilde{a}^l_t, [\hat{\pi}_t]_i, x_{\leq t}) &= [\hat{\pi}_t(x_{\leq t}|\mathrm{do}(A^l_t = \tilde{a}^l_t))]_i - [\hat{\pi}_t(x_{\leq t}|\mathrm{do}(A^l_t = a^l_t))]_i\\\label{eq:ablate-proof2}
    &= [\hat{\pi}_t(x_{\leq t}|\mathrm{do}(A^l_t = \tilde{a}^l_t)) - \hat{\pi}_t(x_{\leq t}|\mathrm{do}(A^l_t = a^l_t))]_i\\\label{eq:ablate-proof3}
    &= [\hat{\pi}_t(x_{\leq t}|\mathrm{do}(A^l_t = \tilde{a}^l_t)) - \hat{\pi}_t(x_{\leq t})]_i\\
    &= \Delta_{\mathrm{ablate}, l}(x_{\leq t}).
\end{align}
where we go from Equation~\ref{eq:ablate-proof2} to~\ref{eq:ablate-proof3} by using the fact that the intervention $\mathrm{do}(A^l_t = a^l_t)$ doesn't change $A^l_t$ from the value it would have taken in the context $x_{\leq t}$, as in Section~\ref{ss:ablation-impact}. This shows that our ablation-based impact measure corresponds to measuring the total effect due to a change from the natural activation distribution to one sampled from the ablation distribution. The total effect of ablation (knockout) measures the importance of a unit in a given inference: if we were to ablate it, how much would performance suffer if the effects of the ablation were to cascade through the network.

\begin{definition}[Direct effect]
The direct effect of altering $Z=z\to Z=z'$ is given by
\begin{equation}
    DE(z\to z', Y, u) = Y\left(u | \mathrm{do}(Z=z', M=m(u))\right) - Y(u | \mathrm{do}(Z=z))
\end{equation}
i.e. the effect of intervening to set $Z=z'$ and then resetting all other variables $M = V\setminus\{Z, Y\}$ to the value they would have obtained in the context $u$. As with the total effect, if $z=z(u)$ then the direct effect reduces to
\begin{equation}
    DE(z\to z', Y, u) = Y\left(u | \mathrm{do}(Z=z', M=m(u))\right) - Y(u).
\end{equation}
\end{definition}
The direct effect measures how much changing the unit's value would affect the output if all other units' values were held constant. Because of this, in a language model only units connected via a residual path to the output (i.e. at the same token position) can have a direct effect - all other units must have their effect mediated by at least one attention head in order to cross between token positions. The residual structure of our language models means that effects on logits are additive (up to a normalisation constant introduced by RMSNorm), and so every change in logits eventually occurs due to a direct effect.

\paragraph{Unembedding-based impact with RMSnorm held constant approximates direct effect} To see the relation between the unembedding-based impact $\Delta_{\mathrm{unembed}, l}$ and the direct effect of an attention variable $A^l_t$ on the logits $\hat{\pi}_t$ in context $x_{\leq t}$ we first rewrite the architecture defined by Equations~\ref{eq:transformer-eqns}-\ref{eq:transformer-eqns-final} in the `unrolled view' introduced by~\cite{veit2016residual} (for more details on this mathematical perspective on transformers see~\citep{elhage2021mathematical}):
\begin{align}
    z^L_t(x_\leq{t}) = \sum^L_{l=1}m^l_t(x_{\leq t}) + a^L_t(x_{\leq t}),
\end{align}
where $L$ is the final layer of the network and we leave the dependence of activations at layer $L$ on earlier layers implicit. To get the logits we simply unembed $z^L_t$:
\begin{align}
    \pi_t(x_{\leq t}) &= \mathrm{RMSNorm}(z^L_t)W_U\\
    &= \frac{z^L_t(x_\leq{t})}{\sigma(z^L_t)}GW_U\\\label{eq:unrolled-network}
    &= \frac{1}{\sigma(z^L_t)}\sum^L_{j=1}\left[m^j_t(x_{\leq t}) + a^j_t(x_{\leq t})\right]GW_U,
\end{align}
where $G$ is the RMSNorm gain matrix and $W_U$ is the unembedding matrix (see Sections~\ref{ss:transformer-architecture} and~\ref{ss:ablation-impact}). Equation~\ref{eq:unrolled-network} demonstrates that the logits (and thus the centred logits $\hat{\pi}$) are linear in the layer outputs so long as the RMSNorm scale factor $\sigma(z^L_t)$ is held constant. Now we can compute the direct effect of ablating attention layer output $a^l_t$ on the logits (we omit centring as in Equation~\ref{eq:logit-centring} here for brevity, but it is trivial to include):
\begin{align}
    DE(a^l_t\to\tilde{a}^l_t, \pi_t, u) &= \left[\pi_t(x_{\leq t} |\mathrm{do}(A^l_t=\tilde{a}^l_t, M=M(x_{\leq t}))) - \pi_t(x_{\leq t})\right]_i\\
    &=\left[\frac{1}{\sigma(z^L_t)}\left(\tilde{a}^l_t + m^l_t(x_{\leq t}) + \sum^{L}_{j\neq l}\left[m^j_t(x_{\leq t}) + a^j_t(x_{\leq t})\right] - \sum^L_{j=1}\left[m^j_t(x_{\leq t}) + a^j_t(x_{\leq t})\right]\right)GW_U\right]_i\\
    &= \left[\frac{\tilde{a}^l_t - a^l_t(x_{\leq t})}{\sigma(z^L_t)}GW_U\right]_i\\
    &= u(\tilde{a}^l_t)_i - u(a^l_t(x_{\leq t}))_i.
\end{align}
The only difference (up to centring) between the direct effect computed above and $\Delta_{\mathrm{unembed}, l}$ is the inclusion of the impact of the ablation on the maximum-likelihood token $u(\tilde{a}^l_t)_i$. This factor is typically negligible if the source of resample ablations are chosen correctly (otherwise the ablation would still be substantially increasing the probability of the maximum-likelihood token) and is zero in the case of zero ablations, in which case $\Delta_{\mathrm{unembed}, l} = DE(a^l_t\to\tilde{a}^l_t, \hat{\pi}_t, u)$.

\begin{definition}[Indirect effect]
The indirect effect of altering $z\to z'$ is given by
\begin{equation}
    IE(z\to z', Y, u) = Y(u|\mathrm{do}(Z=z, M=\tilde{m}) - Y(u|\mathrm{do}(Z=z));\quad \tilde{m}=m(x'_{\leq t})
\end{equation}
which is the effect of setting the variables $M$ to their ablation values while also resetting $Z$ to its default value $z$.
\end{definition}

Indirect effect measures the effect that a unit has via downstream units, i.e. variables that are on the path from $Z$ to $Y$ (we say a variable $M$ satisfying this criterion \textit{mediates} the relationship between $Z$ and $Y$). Units inside a circuit that is important in the current context will have high indirect effect, whereas units at the terminus of the circuit will have high direct effect. When we don't specify the mediating variables we assume that all variables between the intervened variables and the output variables are changing.

\subsection{Challenges and opportunities in intervening in neural networks}
The difficulties and affordances involved in performing causal analysis on neural networks are almost the opposite of those involved in most real-world causal analysis: we know the causal model with complete certainty (down to the level of individual parameters), can perform arbitrarily long chains of interventions rapidly and can read the value of all variables simultaneously. On the other hand, our causal models involved enormous numbers of individual parameters, and these parameters have no obvious meaning attached to them (with the exception of input and output variables). Painstaking analysis often suggests meanings for individual neurons or clusters of neurons~\citep{bau2018gan, carter2019activation, hilton2020understanding, goh2021multimodal} but the correct unit of analysis still remains unclear~\citep{morcos2018importance}. A recent line of work on a phenomenon known as superposition has begun to shed some light on how directions in activation space relate to meaningful units of analysis \citep{elhage2022superposition} but this work has yet to reach any definitive conclusions that allow us to decide how to group neurons to reduce the size of the graph. For this reason we work with ablations at the level of individual layers, while acknowledging that this level of analysis is still likely to be too coarse to capture all the relevant phenomena.

\subsection{Toy model and motif for the Hydra effect}
\begin{figure}
    \centering
    \includegraphics[width=0.7\textwidth]{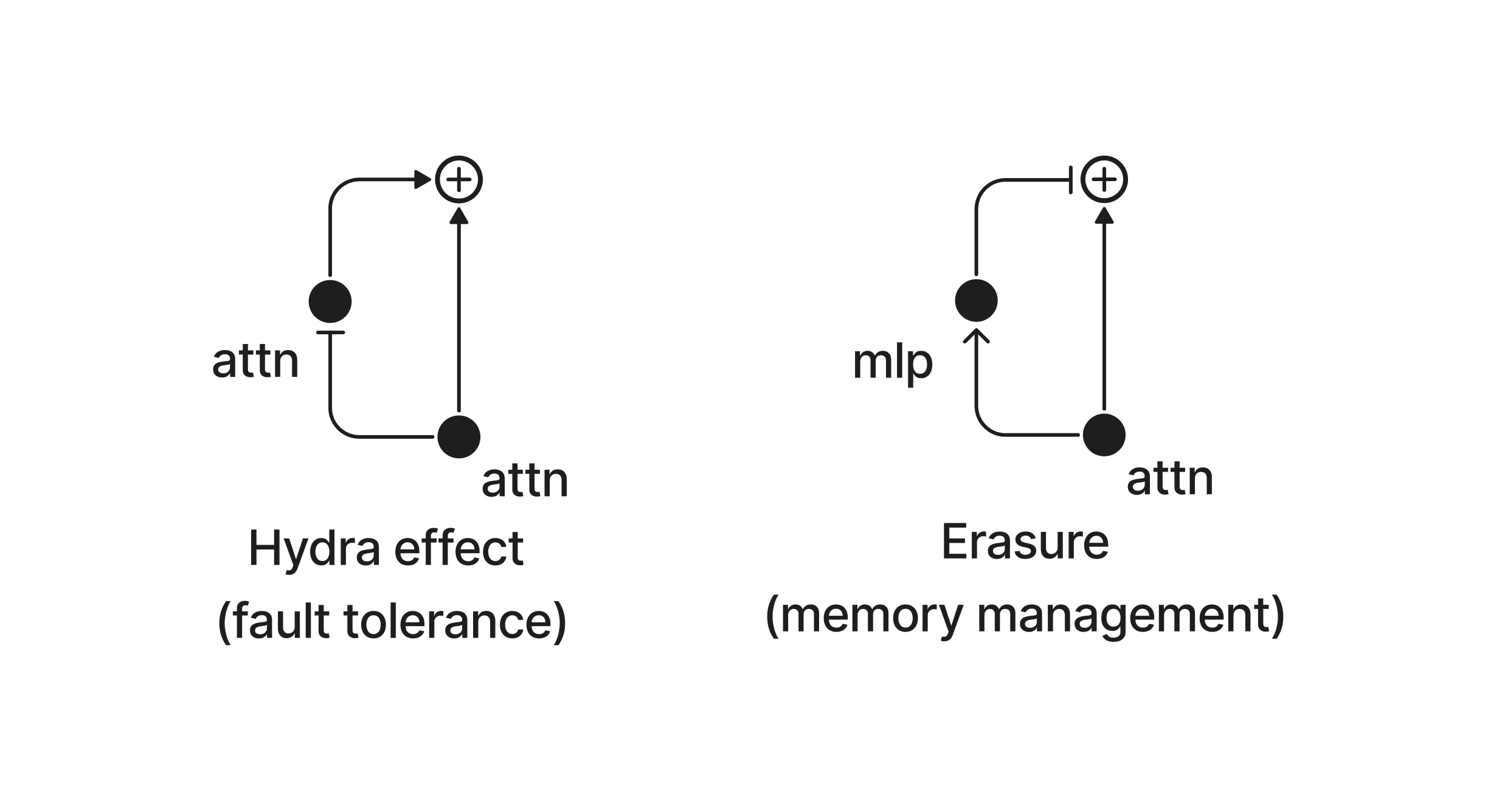}
    \caption{Network motifs that cause total and direct effect to be uncorrelated, either by self-repair or erasure. An arrow-ended line $X\to Y$ indicates that $X$ increases $Y$ whereas a bar-ended line indicates that $X$ inhibits $Y$.}
    \label{fig:motifs}
\end{figure}
We now turn the tools of causal inference that we have just introduced to our findings on self-repair by studying a simplified toy model of the phenomena we observed in Section~\ref{s:hydra-explanation}. These toy models both use a simple causal model involving a single exogenous variable $u$ and endogenous variables $x(u) = u$ and $y$:
\begin{equation}
    y = x(u) + f(x, u).
\end{equation}
If we intervene to set $x=0$, what functions $f$ will ensure that the total effect $TE(x, y)$ is zero? Two simple possibilities stand out:
\begin{equation}\label{eq:erasure}
    f(x, u) = -x\quad\mathrm{(Erasure)}
\end{equation}
which will ensure that $y=0$ regardless of the input variable $u$, or
\begin{equation}
    f(x, u) = \begin{cases}
    0 & \mathrm{if}\,x = u\\
    u & \mathrm{otherwise}
    \end{cases}\quad\text{(Self-repair)}.
\end{equation}
In both cases the output variable $y$ is kept unchanged regardless of the inner workings of the model, but in the erasure case it is clamped to zero whereas in the self-repair case it is fixed at $u$. Although these simple circuits are stylised in order to illustrate a point, they turn out to be surprisingly good models of phenomena we observe in real language models.

\section{Quantifying erasure and the Hydra Effect}\label{s:quantifying-hydra}
\subsection{Methodology}
For a given context $u$ we can measure the total compensatory effect following an ablation $\tilde{a}^m$ by summing the effects of the downstream changes in the network:
\begin{equation}\label{eq:compensatory-effect}
    CE(\tilde{a}^m, u) = \underbrace{\sum^L_{l=m+1}\Delta DE(a^l, u, \tilde{a}^m)}_{\text{Downstream effect on Attns}} + \underbrace{\sum^L_{l=m}\Delta DE(m^l, u, \tilde{a}^m)}_{\text{Downstream effect on MLPs}}\\,
\end{equation}
where $\Delta DE$ is the difference in direct effects between the ablated and unablated networks:
\begin{equation}
    \Delta DE(z^l, u, \tilde{z}^m) = DE_{\text{ablated}}(z^l, u, \tilde{z}^m) - DE(z^l, u),
\end{equation}
where $DE_{\text{ablated}}(z^l, u, \tilde{z}^m) = \tilde{\Delta}^k_{\mathrm{unembed}, l}$ is the direct effect of layer $l$ following an ablation at layer $m$ in context $u$. The starting index of the downstream MLPs and attention layers differs because MLP layers follow attention layers in our model. The compensatory effect for MLP ablations $CE(\tilde{m}^m, u)$ is identical to Equation~\ref{eq:compensatory-effect} except that the MLP sum index starts at $m+1$. We generate a dataset of direct and compensatory effects by computing $DE(a^l, u)$, $DE(m^l, u)$, $CE(a^l, u)$ and $CE(m^l, u)$ for all layers $l$ and all 1,209 contexts $u$ in the Counterfact dataset, i.e. for every prompt we compute the direct and compensatory effects of every possible layer ablation.

\subsection{Results}
\begin{figure}[h!]
    \centering
    \begin{subfigure}{0.45\textwidth}
    \centering
    \includegraphics[width=\textwidth]{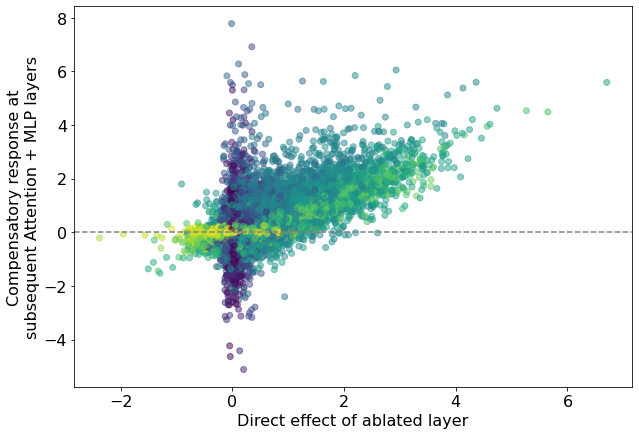}
    \caption{}
    \end{subfigure}
    \begin{subfigure}{0.45\textwidth}
    \centering
    \includegraphics[width=\textwidth]{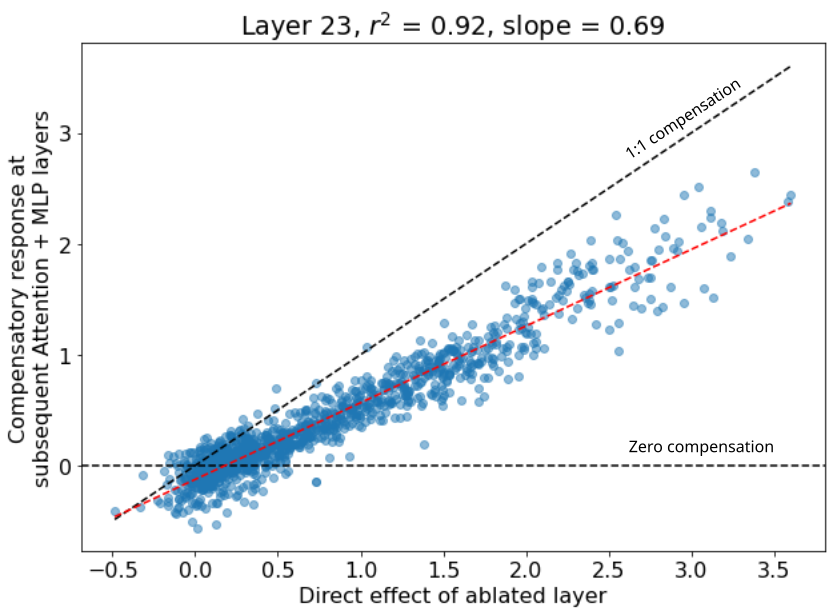}
    \caption{}
    \end{subfigure}
    
    \begin{subfigure}{0.45\textwidth}
    \centering
    \includegraphics[width=\textwidth]{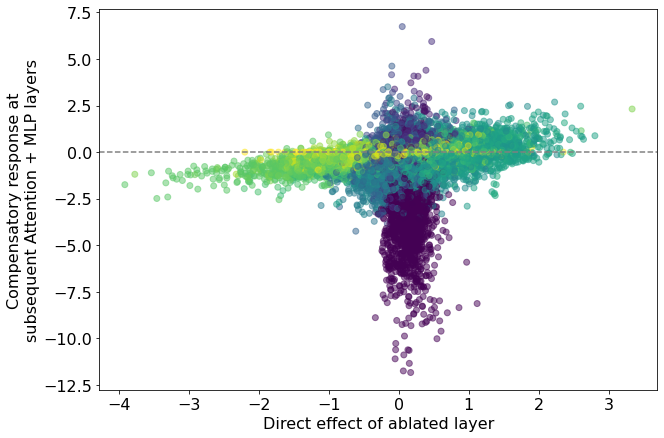}
    \caption{}
    \end{subfigure}
    \begin{subfigure}{0.45\textwidth}
    \centering
    \includegraphics[width=\textwidth]{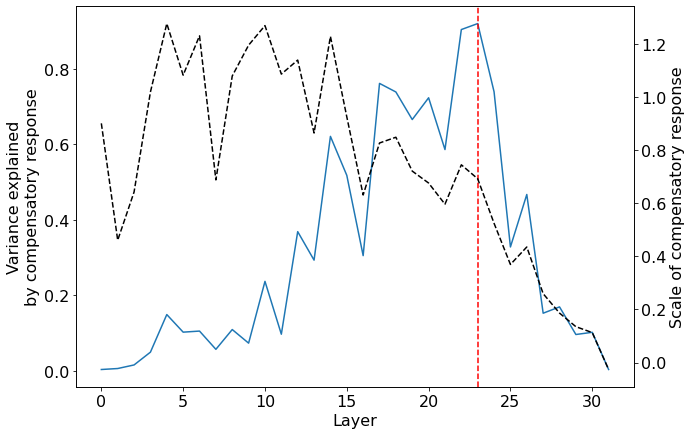}
    \caption{}
    \end{subfigure}
    \caption{Results from carrying out ablations at all layers across the entire Counterfact dataset. (a) Direct effects against compensatory effect for attention layers, with colourmap indicated network depth. Middle layers show strong correlations whereas early and late layers do not. (b) Relation between direct and compensatory effect at the layer with the highest correlation, which occurs at layer 23 where compensation explains 92\% of the variance in changes between the intact and ablated network. (c) Same as (a) but for MLP layers. (d) Summary of variance explained (solid blue line) and slope (dashed black line) of a linear regression between direct and compensatory effects at each attention layer. Red line marks layer shown in subfigure (c).}
    \label{fig:compensation_quantification}
\end{figure}

\begin{figure}[h!]
    \centering
    \includegraphics[width=0.8\textwidth]{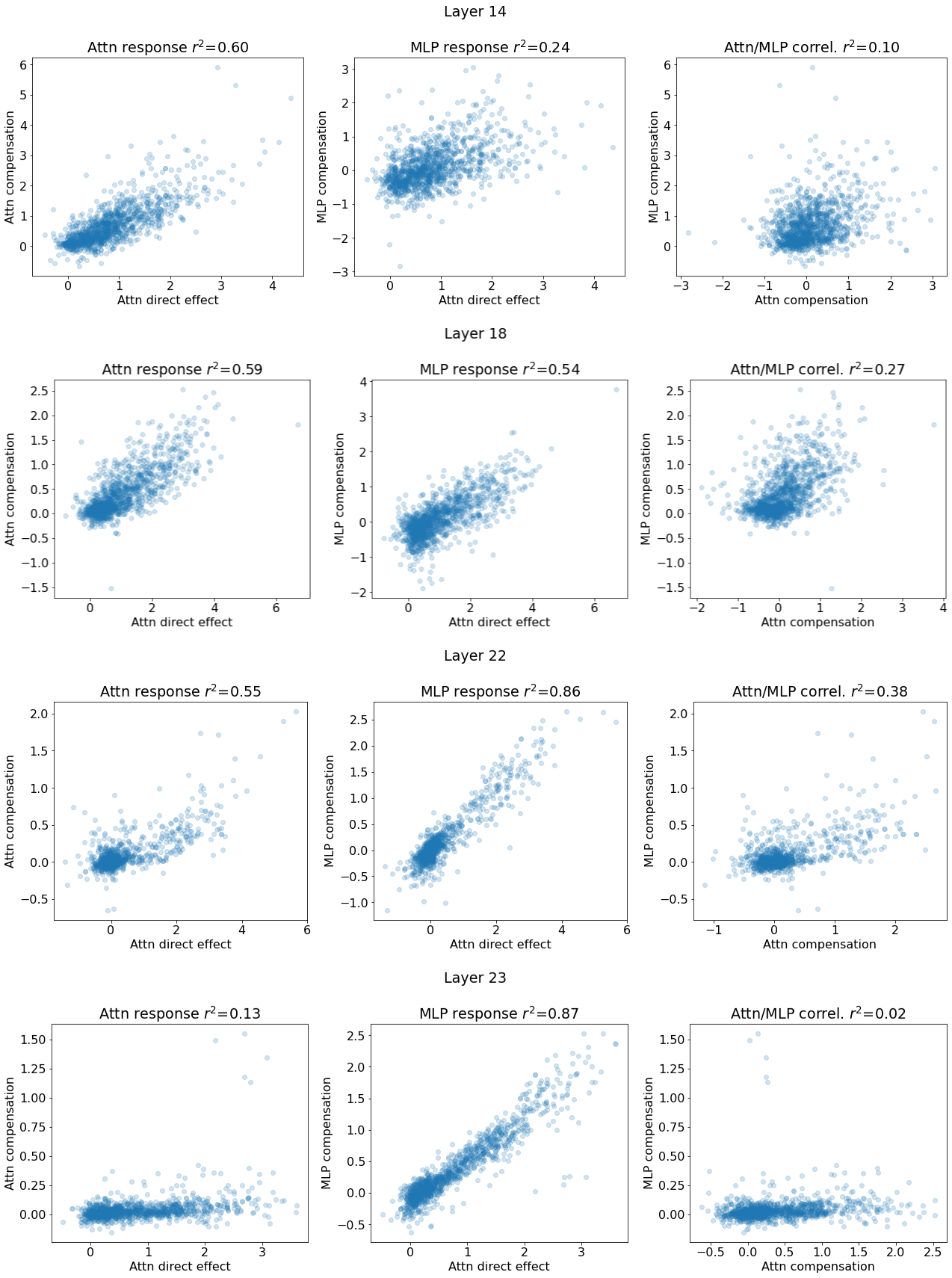}
    \caption{Per-layer compensation results for a selection of ablation layers. The balance between the effect due to the attention and MLP layers shifts from attention to MLP as depth increases.}
    \label{fig:quantification_per_layer}
\end{figure}

Results from full quantification of the compensatory effects across the full Counterfact dataset are shown in Figure~\ref{fig:compensation_quantification}, with data further broken out to individual layers in Figure~\ref{fig:quantification_per_layer}. We highlight the following observations from these results:
\paragraph{Direct and compensatory effects are only well-correlated at intermediate layers:} early layer ablations have large total effects but almost no direct effect (Figure~\ref{fig:quantification_per_layer}a, c, c.f. Figure~\ref{fig:overall-effects}) whereas very late layers only have non-negligible direct effect (which makes sense as there are few downstream layers for them to have an effect on).

\paragraph{The compensatory response is almost entirely responsible for changes in direct effect at later layers.} Compensatory response is very highly correlated at intermediate-late layers (see Figure~\ref{fig:quantification_per_layer}b, d), suggesting that the network's response to ablations at these layers is almost entirely driven by the Hydra effect and a decrease in MLP erasure. Layer 23 (of 32) marks the high point of this phenomenon, with 92\% of the variance in $\Delta DE$ in downstream layers being explained by the Hydra effect.

\paragraph{The compensatory response does not fully restore the output:} fitting a linear regression between direct effect and compensatory response gives a slope of less than one at all layers past layer 13 (see Figure~\ref{fig:compensation_quantification}). This implies that these ablations \textit{do} have a nonzero total effect, but one that is much smaller than it would have been without the Hydra effect and responses from erasure MLPs.

\paragraph{The balance between the Hydra effect and reduction in MLP erasure shifts with network depth} Figure~\ref{fig:quantification_per_layer} shows $\Delta DE$ for attention and MLP layers separately at different layers in the network. In early layers changes in attention direct effect play a considerably larger role, whereas by layer 22 the balance has shifted such that the the MLP response is considerably more predictive and at layer 23 almost all of the response is occuring in the erasure MLPs.

\section{Related Work}\label{s:related-work}
The use of techniques from causal inference to analyse neural networks has been used in a range of cases, including the causal tracing method for locating factual knowledge~\citep{meng2022locating}, mediation analyses for gender bias~\citep{vig2020causal, vig2020investigating} and analysis of syntactic agreement~\citep{finlayson2021causal}. There is also a recent line of work on constructing causal abstractions of neural network computations~\citep{geiger2021causal, geiger2022inducing, massidda2022causal, geiger2023causal, geiger2023finding}. The use of ablations as a way of validating hypotheses about mechanisms in neural networks has been previously suggested~\citep{leavitt2020towards, morcos2018importance, chan2022causal}, although our findings caution against straightforwardly interpreting low effectiveness of ablations as meaning that a network component is unimportant.
\newline\newline
Earlier work on residual networks (of which decoder-only transformers are a special case) determined that for image classification networks, residual networks behave like ensembles of shallower networks~\citep{veit2016residual}. This work introduced both the `unravelled view' of residual networks that we make use of and experimented with ablating network layers at test time, determining that most effective paths through residual networks are short and layers often do not depend on one another.
\newline\newline
The idea of interpreting neural networks in terms of their internal mechanisms or circuits~\cite{olah2020zoom} (often referred to as mechanistic interpretability) is relatively recent. Earlier work on vision models~\citep{olah2018building} identified human-understandable neurons and simple circuits (for instance curve detectors~\citep{cammarata2020curve}). Subsequent work on transformers has identified `induction circuits' responsible for simple instances of in-context learning~\citep{elhage2021mathematical, olsson2022context}, as well as a mechanism for indirect object identification~\citep{wang2022interpretability} and the mechanism underlying the `grokking' phenomenon~\citep{power2022grokking, nanda2023progress, chughtai2023toy}.
\newline\newline
The use of probes to analyse neural network training originated in~\citep{alain2016understanding}, and has been widely used since. In the context of transformer language models the `logit lens' approach, which involves using the model's own unembedding matrix to decode the residual stream, has been applied to early GPT models~\citep{nostalgebraist2020logit}. In order to better align with the model's true outputs~\cite{belrose2023eliciting} use a learned affine unembedding rather than the model's unembedding matrix and are also able to perform causal interventions using a learned `causal basis'. \cite{geva2022transformer} and follow-on work~\citep{geva2022lm, dar2022analyzing} analyse MLP layers by unembedding specific subspaces of their outputs. Sparse probing has been used to empirically study the superposition phenomenon~\citep{elhage2022superposition} in large language models~\citep{gurnee2023finding} and has been used to understand concept learning in deep reinforcement learning agents~\citep{mcgrath2022acquisition, forde2022concepts}.



\section{Conclusion}\label{s:conclusion}
\paragraph{Findings} In this work we have investigated the computational structure of language models during factual recall by performing detailed ablation studies. We found that networks exhibit surprising self-repair properties: knockout of an attention layer causes another attention layer to increase its effect in order to compensate. We term this new motif the Hydra effect. We also found that late-layer MLPs appear to have a negative-feedback/erasure effect: late layer MLPs often act to reduce the probability of the maximum-likelihood token, and this reduction is attenuated when attention layers promoting that token are knocked out. We find that these effects are approximately linear, and that at middle layers (where these effects are strongest) the Hydra effect and reduction in MLP effects collectively act to restore approximately 70\% of the reduction in token logits.

\paragraph{Implications for interpretability research} These findings corroborate earlier work on neural network computations in GPT-2 Small~\citep{wang2022interpretability} which reported a similar effect that the authors term `backup heads'. The authors of~\citep{wang2022interpretability} hypothesised that dropout~\citep{srivastava2014dropout} was responsible for self-repair behaviour, which we disprove as the model we study (Chinchilla 7B) was trained without any form of dropout or stochastic depth. The occurrence of this motif across tasks and models suggests that it may be an instance of universality~\citep{olah2020zoom}. Our original motivation for this work was performing automated ablation studies using an algorithm similar to~\citep{conmy2023towards}, which led to us investigating the changes in network computation under repeated ablations. The Hydra effect poses a challenge to automating ablations: if we prioritise network components for ablation according to their total effect, we will be using a measure that does not fully reflect the computational structure of the intact network. Fortunately, the fact that the compensatory effect is typically less than 100\% means that automated ablations will still have some signal to work with. The Hydra effect and erasure MLPs also have implications for attributing responsibility for a network's output to individual network components: is the responsible component the attention layer that has the effect in the intact network, or the circuits that act to compensate following ablation? The framework of actual causality~\citep{halpern2016actual} may be a useful way to approach this question. 

Our findings also suggest that attempting to assign semantics to MLP neurons may be more complicated than otherwise expected: erasure MLPs may have no clear semantics in terms of the model's input, as they are responding to the language model's internal computations. Finally, our findings also have implications for work that attempts to understand language models by unembedding the output of individual layers (e.g.~\citep{geva2022transformer}) - this corresponds to an assumption that the direct effect is the only meaningful effect of a layer. The existence of erasure MLPs poses a challenge to this approach to interpretability: if the output of an attention layer or MLP is guaranteed to be partially undone by an erasure MLP, it's no longer straightforward to interpret that output in terms of its direct effects on logits: the effect of the mediating MLPs should also be considered. Our findings also provide context for earlier ablation studies (such as~\citep{morcos2018importance}): it is not enough simply to measure the total effect of an ablation without investigating downstream changes in the network, and more important network components are more likely to be robust to ablation.

\paragraph{Implications for language modelling research} From the perspective of language modelling the Hydra effect is surprising: it confers robustness to a kind of ablation that the model will never experience at inference time and so appears to be a waste of parameters. If this is the case, what benefit does it confer? One possibility (drawing on the analytical framework of Tinbergen's four questions~\citep{tinbergen1963aims}) is that the Hydra effect confers no benefit at inference time, but is beneficial in the context of training. If gradient descent were to occasionally break network components then a kind of `natural dropout' would occur during training. In this case it would be beneficial for networks to be robust to layers failing. We emphasise that this is conjecture, however, and would need further research.

\paragraph{Possible extensions}
Although we have identified two new motifs, we have not investigated more deeply than individual layers (for instance looking at the level of individual attention heads or directions in activation space). Achieving a greater level of precision is a natural next step and would unlock deeper understanding of the mechanisms at play here. Some questions that could be answered with a finer-grained understanding of how this kind of redundancy operates include:
\begin{enumerate}
    \item How much does the Hydra effect occur across the entire training distribution? Does sequence length play any role?
    \item What are the Hydra effect attention heads responding to the presence/absence of in the residual stream?
    \item Do the same downstream heads act as Hydra effect replacement heads across multiple contexts?
    \item What causes the Hydra effect? Is the natural dropout hypothesis correct or is some other phenomenon responsible (superposition has been suggested as an alternative explanation).
    \item Is there a Hydra effect for features rather than direct effect on logits?
    \item What are the erasure heads responding to in the residual stream? Do they have a `recalibration' effect when a wider range of tokens is considered?
    \item If we account for redundancy/Hydra effect, can we probe network structure by using targeted ablations?
\end{enumerate}

\paragraph{Acknowledgements}
We would like to thank Joe Halpern, Avraham Ruderman, Tom Lieberum, Chris Olah, Zhengdong Wang, Tim Genewein and Neel Nanda for helpful discussions.

\bibliography{main}

\appendix
\section{Choice of intervention distribution}\label{s:appendix-intervention-distribution}
When we intervene on a neural network's activations with the $\mathrm{do}$-operator, we are setting some subset of these activations to a new value in the forward pass and allowing the effect of these changes to propagate forward. Both the corrupted run and the patching operation in the causal tracing method~\citep{meng2022locating} are examples of interventions. Practically we can accomplish these interventions via PyTorch hooks, JAX's Harvest functionality, or passing values for intervened-upon tensors in Tensorflow's feed-dict. When we intervene to set some network activation $Z$ to an `ablation' value $\tilde{z}$, what value should we use? Four main possibilities have been suggested:
\begin{align}
    \textbf{Zero ablation}&: \tilde{z} = 0,\\
    \textbf{Mean ablation}&: \tilde{z} = \mathbb{E}_{u\sim P(u)}\left[Z(u)\right],\\
    \textbf{Noise ablation}&: \tilde{z} = Z(u + \epsilon),\,\epsilon\sim\mathcal{N}(0, \sigma^2),\\
    \textbf{Resample ablation}&: \tilde{z} = Z(\tilde{u}),\, \tilde{u}\sim P(u).
\end{align}
Of these, zero-ablation is the simplest to implement but is typically out-of-distribution for networks trained without some form of layer dropout or stochastic depth. Noise ablation was used extensively in causal tracing~\citep{meng2022locating}. Resample ablation (as introduced by~\cite{chan2022causal}) is more complicated to implement but is probably the most principled, as every ablation is a naturally-occurring set of activations, and so is more likely to respect properties of network activations such as emergent outliers~\citep{dettmers2022llm}. Resample ablation also has the appealing property that by specifying the distribution of inputs $P(u)$ we can control for properties of the input that might otherwise confound our results. To get meaningful results from sample ablations it is necessary to use an average of sample ablations across multiple samples from the same dataset, i.e. a Monte-Carlo approximation to:
\begin{equation}
    V_{Z}(u) = \int V(u | do(Z=\tilde{z}))\, p(\tilde{z})\, d\tilde{z},
\end{equation}
where $p(\tilde{z}) = \int p(Z(u)) du$ is the probability of getting activation values $\tilde{z}$. Note that mean ablation and resample ablation are quite different: mean ablation ablates with the \textit{average activation}, whereas resample activation averages the \textit{effects} of different ablations. See~\citep{chan2022causal} for an extended discussion of these methodological details.

\end{document}